Towards Realization of Augmented Intelligence in Dermatology: Advances and Future Directions

Roxana Daneshjou[1], Carrie Kovarik[2,3], and Justin M Ko[1,3]

1. Department of Dermatology, Stanford School of Medicine, Redwood City, CA 94063
2. Department of Dermatology, University of Pennsylvania, Philadelphia, Pennsylvania
3. American Academy of Dermatology Task Force on Augmented Intelligence

Artificial intelligence (AI) algorithms using deep learning have advanced the classification of skin disease images; however, these algorithms have been mostly applied "in silico" and not validated clinically. Most dermatology AI algorithms perform binary classification tasks (e.g. malignancy versus benign lesions), but this task is not representative of dermatologists' diagnostic range.[1,2] The American Academy of Dermatology Task Force on Augmented Intelligence published a position statement emphasizing the importance of clinical validation to create human-computer synergy, termed augmented intelligence (AuI).[3]

Liu et al's recent paper, "A deep learning system for differential diagnosis of skin diseases" represents a significant advancement of AI in dermatology, bringing it closer to clinical impact.[4] They leveraged a real-world teledermatology dataset to develop an algorithm combining clinical data (demographic information, signs/symptoms, medical history) with multiple clinical photos to generate a differential across 26 diagnoses. They evaluated the algorithm's ability to differentiate between clinically relevant categories such as malignancy versus non-malignancy and infectious versus non-infectious. Liu et al compared the performance of the algorithm against nurse practitioners, primary care physicians and dermatologists, the former two groups being outperformed by the algorithm. They suggest that deploying the algorithm to non-specialist clinicians may aid in generating broader and more accurate differential, potentially improving clinical decision-making.

Liu et al's work brings AI one step closer to the goal of AuI; however, significant issues must be addressed before this algorithm can be integrated into clinical workflow.  These issues include accurate and equitable model development, defining and assessing appropriate clinical outcomes, and real-world integration.

A successful clinical AI algorithm requires training and testing datasets that can produce a generalizable, action-informed, clinically-impactful algorithm.[5]  While the real-world teledermatology data used by Liu et al is superior to textbook photos, we do not know if teledermatology data is generalizable to in-person care.  Types of cases assessed by teledermatology may differ from those seen in-person.  Additionally, structural aspects of teledermatology programs may introduce biases.  Prospective testing in a clinical setting are a crucial next step to validate in a real world environment.

The accuracy of labeled data is another important consideration.  Dermatologists not only rely on the clinical exam and history to make a diagnosis, but also additional diagnostic tools – wound cultures, potassium hydroxide preps, lab work, dermoscopy, and biopsies.  Even for clinical diagnoses, such as psoriasis, the gold standard is an in-person visit with a full skin exam, history, and necessary diagnostic workup. For skin cancers, the gold standard is clinical-pathological diagnosis after biopsy.  Liu et al relied predominantly on consensus diagnosis of images by board-certified dermatologists.[4]  However, not all the dermatologists who labeled the images practiced teledermatology, and experience in clinic may not translate to teledermatology.[4]  Further, the inter-rater variation in diagnosis between clinicians can be significant, and assessment of dermatologist accuracy against the algorithm may be reflective of this heterogeneity.  Only a small subset of cases in Liu et al's paper had biopsy results; in the validation data used to compare the algorithm against clinicians, only 52 cutaneous malignancies were biopsy-proven (6 were melanoma).[4]  Pigmented lesions are especially difficult to diagnose based on consensus alone – a meta-analysis found that among dermatologists, the number needed to biopsy to identify one melanoma was 9.6 for general dermatologists and 5.85 for pigmented lesion specialists.[6]  Amelanotic melanomas are even more difficult to diagnose, given their atypical appearance, leading to delayed diagnosis and worse outcomes.[7]  Ideally, AI algorithms training on skin cancer images should have pathological confirmation.  In lieu of this, any skin cancer detecting algorithm must be validated on pathologically confirmed skin cancers, preferably through a prospective clinical trial.

While a critical step in the diagnostic process, algorithmic generation of differential diagnosis is still removed from clinical decision-making. Physicians worry about the "must not miss" diagnosis – an uncommon diagnosis that can be life altering if not diagnosed and treated. Examples in dermatology include life threatening drug reactions, such as Stevens' Johnson/Toxic Epidermal Necrosis (TEN) and drug reactions with eosinophilia and systemic symptoms (DRESS). Other examples include rare but aggressive forms of cancer, such as Merkel cell carcinoma or angiosarcoma. Because of their rarity, these examples were not represented in Liu et al's algorithm. Even common diagnoses such as squamous cell carcinoma may have an atypical presentation, be more aggressive, and become a "must not miss" diagnosis, such as with severely immunocompromised patients. Because of the severe consequences, deployment of an AI algorithm, particularly if used for triage by non-specialists, must have a mechanism, or human in the loop, to allow for identification of "must not miss" conditions.

Significant health disparities persist in medicine, and dermatology is no exception. Patients with skin of color present with melanoma at later stages and have worse outcomes.[8] Skin disease presents differently across different skin tones, yet there are gaps in representation of diverse skin tones in dermatology textbooks.[9] In Liu et al's datasets, Fitzpatrick skin type (FST) V/VI, the two darkest skin types, were considerably underrepresented. Validation data had less than 3% FST skin type V cases and only a single case with FST skin type VI.[4] Most published dermatology AI algorithms are not trained or tested on diverse datasets.[10] AI algorithms in dermatology should demonstrate robust performance across diverse skin tones, so that technology bridges rather than exacerbates existing biases and structural inequities.

When creating datasets to be used for AI algorithms, we must also consider privacy and consent for use of patient data. Dermatology images, though often de-identified, are still personal in nature. Companies developing AI algorithms can purchase skin disease photos from healthcare practices, and patients are often unaware and do not give explicit consent.[11] Once an image has been used to create an algorithm, it becomes difficult to remove if a patient rescinds consent.[3] In the European Union, the General Data Protection and Regulation of May 2018 requires that patients provide explicit informed consent before any personal data is collected; this data must also be "trackable" and removed upon patient request.[11] In the United States, radiologists are raising the importance of data stewardship and appropriate consenting in AI clinical research.[12]

The ultimate goal is AuI algorithms that enable delivery of higher value collaborative care. In dermatology, these algorithms aim to assist the patient or non-specialist (triage and referral decisions) or the dermatologist (decision making and diagnosis). In either setting, AI algorithms must be validated in prospective clinical trials and demonstrate impact on relevant clinical outcomes. Example outcomes include shortening time to diagnosis for malignant neoplasms or better identifying patients appropriate for dermatology referral from the primary care setting. Additionally, we must evaluate how the interaction between algorithms and clinicians influence clinical decision-making, as previous studies show that a priori clinician confidence can affect whether AI alters decision making.[13] Finally, we must understand patient perceptions toward AI so that we can deliver integrated care that is sensitive to patient values and preferences.[11,14]

AI in dermatology is a rapidly evolving field with the potential to transform how we deliver care through AuI – the harnessing of synergy between clinicians and machine. Liu et al's work simultaneously propels the field forward, while unveiling a path that needs to be traversed safely, effectively, and equitably in order to wield these tools for patient benefit.